\documentclass[11pt]{article}

\usepackage[]{EMNLP2023}

\usepackage{times}
\usepackage{latexsym}

\usepackage[T1]{fontenc}

\usepackage[utf8]{inputenc}

\usepackage{microtype}

\usepackage{inconsolata}
\usepackage{enumitem}
\usepackage{multirow}
\usepackage[inline]{trackchanges}
\addeditor{YJ}
\addeditor{WD}

\usepackage{amsmath,amsfonts,amssymb,bm}
\usepackage{pifont} 
\usepackage{graphicx,epsfig,fancybox} 
\usepackage{float}
\usepackage{color} 
\usepackage[ruled,vlined]{algorithm2e} %
\usepackage{algorithmic}
\usepackage{multirow}
\usepackage{comment}
\usepackage{hyperref}
\usepackage{natbib} 
\usepackage{subfigure}

\renewcommand{\vec}[1]{\boldsymbol{#1}}

\usepackage{adjustbox}
\usepackage{array}
\usepackage{booktabs}

\newcolumntype{R}[2]{%
    >{\adjustbox{angle=#1,lap=\width-(#2)}\bgroup}%
    l%
    <{\egroup}%
}

\title{Blending Reward Functions via Few Expert Demonstrations for Faithful and Accurate Knowledge-Grounded Dialogue Generation}

\author{
Wanyu Du \quad Yangfeng Ji\\
Department of Computer Science\\
University of Virginia\\
Charlottesville, VA 22903\\
\texttt{\{wd5jq,yangfeng\}@virginia.edu}\\
  }

\begin{document}
\maketitle

\begin{abstract}

The development of trustworthy conversational information-seeking systems relies on dialogue models that can generate faithful and accurate responses based on relevant knowledge texts. 
However, two main challenges hinder this task. 
Firstly, language models may generate hallucinations due to data biases present in their pretraining corpus. 
Secondly, knowledge texts often contain redundant and irrelevant information that distracts the model's attention from the relevant text span.
Previous works use additional data annotations on the knowledge texts to learn a knowledge identification module in order to bypass irrelevant information, but collecting such high-quality span annotations can be costly.
In this work, we leverage reinforcement learning algorithms to overcome the above challenges by introducing a novel reward function.
Our reward function combines an accuracy metric and a faithfulness metric to provide a balanced quality judgment of generated responses, which can be used as a cost-effective approximation to a human preference reward model when only a few preference annotations are available.
Empirical experiments on two conversational information-seeking datasets demonstrate that our method can compete with other strong supervised learning baselines.

\end{abstract}
\section{Introduction}
\begin{table*}[t]
    \centering
    \small
    \begin{tabular}{p{0.15\linewidth}|p{0.75\linewidth}}
    \toprule
    \textbf{History} & 
    User: What kind of rockets does SpaceX have?
    ||Agent: One is called the Falcon Heavy and can put about 8,000 kilograms into geostationary transfer orbit with side boosters. 
    ||User: Wow, that's pretty cool. Do you know anything about space elevators?
    \\
    \midrule
    \textbf{Knowledge Text} &  \textcolor{teal}{The design would permit vehicles to travel along the cable from a planetary surface, such as the Earth's, directly into space or orbit, without the use of large rockets.} \\
    \midrule
    \textbf{Human Reference} & Yeah, they would allow vehicles to move along a cable into space, instead of on a rocket. \\
    \midrule
    \textbf{LLM Output} &  Yes, a space elevator is a proposed type of transportation system. It consists of a cable anchored to the surface of a planet, with the other end in space. \textcolor{gray}{\textit{The cable would be held up by the centrifugal force generated by its orbital motion around. (unfaithful)}} \\
    \bottomrule
    \toprule
    \textbf{History} & User: How do i apply for Medicare? \\
    \midrule
    \textbf{Knowledge Text} & Benefits Planner: Retirement | Applying for Medicare Only | Social Security Administration // Applying for Medicare Only.
    If you are within three months of age 65 or older and not ready to start your monthly Social Security benefits yet, \textcolor{teal}{you can use our online retirement application to sign up just for Medicare and wait to apply for your retirement or spouse's benefits later}.  
    Before you decide, you need to 
    \underline{be sure that you} \underline{understand how waiting until later will affect: the lifetime benefits we can pay on your account}
    \underline{and your health insurance coverage}.\\
    \midrule
    \textbf{Human Reference} & You can use our online retirement application to sign up just for Medicare and wait to apply for your retirement or spouse's benefits. \\
    \midrule
    \textbf{LLM Output} &  You can use the Social Security Administration's online retirement application to sign up for Medicare only. \textcolor{red}{Be aware that if you wait to apply for retirement or spouse benefits later, it may affect the benefits you receive and your health insurance coverage. (semantically incorrect)} \\
    \bottomrule
    \end{tabular}
    \caption{\label{tab:intro_example}
    Two example responses generated by a LLM (text-davinci-003) in knowledge-grounded information-seeking conversations. 
    The \textcolor{teal}{green text} highlights the ground-truth knowledge span.
    The \textcolor{gray}{\textit{gray text}} is the unfaithful model-generated response caused by the bias from the LLM. 
    The \textcolor{red}{red text} is the incorrect model-generated response misled by the \underline{text} that contains redundant information.
    }
\end{table*}

Recent large language models (LLMs) have enabled conversational information-seeking systems to exhibit remarkable proficiency in producing fluent and coherent responses \citep{thoppilan2022lamda,nakano2022webgpt,menick2022teaching,ouyang2022training}. 
However, the models sometimes fail in generating faithful and accurate responses supported by verified knowledge texts.
This undesirable model behavior stems from three distinct sources: the bias inherent in LLMs, the irrelevant information in input knowledge texts, and the characteristics of the employed learning algorithms.
Firstly, LLMs are likely to generate texts that are most frequently seen during pretraining \citep{kandpal2022large,wang2023selfinstruct}, and may either disregard the knowledge texts or generate additional information not provided in the knowledge texts, as illustrated in \autoref{tab:intro_example}. 
This problem is also described as hallucinations in other works \citep{maynez-etal-2020-faithfulness,cao-etal-2022-hallucinated,10.1162/tacl_a_00529}. 
Secondly, the input knowledge texts often contain redundant and irrelevant information that diverts the model's attention away from the relevant text span \citep{chen-etal-2021-unsupervised}, as demonstrated in \autoref{tab:intro_example}. 
Lastly, the model is trained to maximize the likelihood of ground-truth responses, which is easy to suffer from exposure bias \citep{ranzato2016sequence,du-ji-2019-empirical}, and sometimes fails to align with the factual content of the input knowledge texts \citep{wang-sennrich-2020-exposure,li2022faithfulness}.

Knowledge-grounded dialogue generation techniques are commonly leveraged in building conversational information-seeking systems.
Aligning with the problem source analysis, we classify the related prior works into three categories.
{The first category of works} proposes to use additional data annotations on the knowledge texts to filter out redundant and irrelevant information, such as learning a knowledge selection module \citep{dinan2018wizard,feng-etal-2021-multidoc2dial,Kim2020Sequential,wu-etal-2021-dialki}, or leveraging cross-attention scores to identify important texts \citep{Zhao2020Low-Resource,xu2021attentionguided}.
While directly learning from additional data annotations can improve the accuracy of locating the intended text spans, high-quality annotated data is often expensive to collect. 
{The second group of works modifies} the model architecture \citep{prabhumoye-etal-2021-focused} or {refines} the decoding process \citep{tian2020sticking} to force the model generating responses more faithful to the knowledge texts.  
{The third line of works switches to} alternative learning algorithms to explore the optimal dialogue generation policy, such as unlikelihood training \citep{li-etal-2020-dont}, imitation learning \citep{liu-etal-2018-dialogue} and reinforcement learning \citep{li-etal-2016-deep,NEURIPS2020_1f89885d,ramamurthy2023is}.
However, previous learning algorithms focus on optimizing semantic coherence and fluency of generated responses, and it is still less studied in improving the faithfulness and accuracy of generated responses in knowledge-grounded conversations.

In this work, we apply reinforcement learning (RL) algorithms to learn faithful and accurate dialogue generation policy. 
On one hand, fine-tuning LLMs with RL on the downstream datasets can help alleviate the bias learned from the pretraining corpus; 
on the other hand, an appropriate reward function can guide LLMs to generate responses that align with the relevant knowledge text.
Our {key contribution} is the design of a novel reward function that {combines} two automatic metrics {via expert demonstrations} for effective evaluation.
{In this design, each} metric {aims to address one specific concern of response generation discussed above}: 
(1) the accuracy metric measures the similarity between model-generated responses and ground-truth references, which aims at making the generated response coherent with the dialogue context; 
(2) the faithfulness metric evaluates the similarity between model-generated responses and input knowledge texts, which aims at aligning the factual content of the generated response with the knowledge text.

We blend the two automatic metrics to approximate a balanced quality judgment of generated responses. 
The blending coefficient is learned from a few expert demonstrations of pair-wise quality judgment on two LLMs' outputs.
Our reward function can be used as a cost-effective approximation to a human preference reward model \citep{NEURIPS2020_1f89885d}, when there only exists a few (e.g. 25) human preference annotations. 
This approach enables the improvement of both faithfulness and accuracy in knowledge-grounded dialogue generation, while reducing the reliance on massive human preference annotations. 
Empirical experiments on two information-seeking conversation benchmark datasets, MultiDoc2Dial \citep{feng-etal-2021-multidoc2dial} and FaithDial \citep{10.1162/tacl_a_00529}, show that our method can obtain improved performance in faithfulness and accuracy compared with other strong supervised learning baselines.

We summarize the contributions of this work as follows:
\begin{enumerate}[noitemsep,topsep=0pt,leftmargin=*]
    \item Identifying three major sources for the problem of generating unfaithful and inaccurate responses in the knowledge-grounded conversations.
    \item Proposing a new reward function for reinforcement learning algorithms that can improve the faithfulness and accuracy of generated responses.
    \item Conducting empirical experiments to demonstrate the effectiveness of our method compared with strong supervised learning baselines.
\end{enumerate}
\section{Related Work}


\paragraph{Knowledge-Grounded Dialogue Generation.}
Previous works in knowledge-grounded dialogue generation train language models conditioning on knowledge texts, with the goal of maximizing the likelihood of ground-truth responses \citep{Ghazvininejad_Brockett_Chang_Dolan_Gao_Yih_Galley_2018,dinan2018wizard,zhang-etal-2020-dialogpt,bao-etal-2022-plato,peng2022godel}. 
Some works apply multi-task learning and transfer learning techniques to improve dialogue generation quality via joint learning with other text generation tasks \citep{shuster-etal-2020-dialogue,JMLR:v21:20-074,li-etal-2022-knowledge}. 
In addition, \citet{zhan-etal-2021-augmenting} learns the knowledge transition in multi-turn conversations to better select knowledge texts for response generation. 
Other works leverage the retriever-reader architecture to learn knowledge text  representations for improving generation quality \citep{NEURIPS2020_6b493230,izacard-grave-2021-leveraging}.
This work leverages reinforcement learning algorithms to learn faithful and accurate dialogue generation policy.

\paragraph{Knowledge Identification in Document-Grounded Dialogues.}
A majority of works incorporate a knowledge identification module in the document-grounded dialogue generation task \citep{dinan2018wizard,Kim2020Sequential,chen-etal-2020-bridging,feng-etal-2021-multidoc2dial}.
The knowledge identification module prevents the language model from attending to irrelevant knowledge texts and avoids generating inappropriate responses \citep{chen-etal-2020-bridging}.
One line of works learns a knowledge identification module with explicit span annotations before generation \citep{wu-etal-2021-dialki,zhao-etal-2020-knowledge-grounded}.
Another line of works models the grounding knowledge texts as latent variables \citep{Zhao2020Low-Resource,Kim2020Sequential}.
This work does not require data annotations on knowledge texts nor applied modification to the model architecture.
We fine-tune LLMs with RL using our novel reward function to get improved performance in accuracy and faithfulness.

\paragraph{Faithful Text Generation.}
The faithfulness text generation problem is defined as whether the generated content is factually consistent with the input information \citep{li2022faithfulness}.
Some works improve the factual consistency of dialogues with a natural language inference model to select the most faithful candidates during inference \citep{welleck-etal-2019-dialogue,qin-etal-2021-dont}, or use the entailment score from the inference model as a reward to learn better dialogue policy \citep{Song_Zhang_Hu_Liu_2020,mesgar-etal-2021-improving}. 
\citet{tian2020sticking} design constrained decoding strategies to improve the faithfulness.
\citet{nye2021improving} propose a dual-system, where the first system generates a set of candidate responses and the second system validates if the generated responses contain contradictions or commonsense violations.
Our method combines the faithfulness and accuracy metrics to better approximate human preference in the low-data setting.
\section{Learning Faithful and Accurate Generation Policy with RL}


\subsection{Problem Definition}
Given the knowledge text $K_n$ and the conversation history $X=(\vec{u}_0, \cdots, \vec{u}_{n-1})$, the task is to generate a response $\vec{u}_n$ that is faithful to $K_n$ and coherent to the conversation history $X$.
As the coherence evaluation in dialogue generation varies across different specific tasks and domains, in this work, we follow prior works \citep{feng-etal-2021-multidoc2dial,10.1162/tacl_a_00529} and approximate it by calculating the accuracy between generated response $\vec{u}_n$ and ground-truth reference $\vec{y}_n$.

Following \citep{10.5555/3312046,ramamurthy2023is}, we formulate the response generation $\vec{u}_n=(a_0, \cdots, a_T)$ as a Markov Decision Process (MDP) $\langle\mathcal{S}, \mathcal{A}, \mathcal{P}, \mathcal{R}, \gamma\rangle$.
$\mathcal{S}$ is a finite set of states, where the initial state $s_0\in\mathcal{S}$ is a concatenation of input conversation history $X$ and knowledge text $K_n$.
$\mathcal{A}$ is a finite set of actions, where an action $a_t\in\mathcal{A}$ is a token from our vocabulary $\mathcal{V}$.
$\mathcal{P} : \mathcal{S} \times \mathcal{A} \to \mathcal{S}$ is a transition function that determines the next state $s_{t+1}$ given the current state action pair $(s_t, a_t)$.
$\mathcal{R} : \mathcal{S} \times \mathcal{A} \to \mathbb{R}$ is a reward function that returns a real number given the current state action pair $(s_t, a_t)$.
$\gamma \in [0, 1]$ is a discount factor.
Each episode in the MDP begins by sampling a datapoint $(\vec{u}_0, \cdots, \vec{u}_{n-1}, \vec{y}_n, K_n)$ from the dataset, and ends when the current time step $t$ exceeds the horizon $T$ or an end of sentence token is generated.

\subsection{Proximal Policy Optimization (PPO)}
The policy $\pi_\theta : \mathcal{S} \to \mathcal{A}$ is a function that selects an action in a given state in order to maximize the long-term discounted rewards over a trajectory $\mathbb{E}_{\pi}[\sum_{t=0}^T \gamma^t R_t]$.
In this work, we initialize the policy $\pi_\theta$ with a pre-trained language model $\pi_0$.
We learn the policy using the Proximal Policy Optimization (PPO) algorithm \citep{schulman2017proximal}, which is an effective actor-critic algorithm in many text generation tasks \citep{NEURIPS2020_1f89885d,ramamurthy2023is}.
The advantage is approximated using Generalized Advantage Estimation \citep{schulman2018highdimensional}: $A_t = \sum_{\tau=t}^T(\gamma\lambda)^{\tau - t}(R_\tau + \gamma V_\phi(s_{\tau+1})-V_\phi(s_{\tau}))$, where $\lambda$ is a hyper-parameter, $\gamma$ is the discount factor, $R_t$ is the reward assigned to $a_t$, and $V_\phi(s_t)$ is the value of state $s_t$ given by the value network $V_\phi$.
Note that the reward $R_t$ is regularized using a token-level KL penalty \citep{wu2021recursively}, in order to prevent the updated policy $\pi_\theta$ deviating too far from the pre-trained language model $\pi_0$:
\begin{equation}
\label{eq:ppo_reward}
    R_t = R_t -\beta \text{KL}(\pi_\theta(a_t|s_t)|| \pi_0(a_t|s_t))
\end{equation}
where $\beta$ is a dynamically adapted coefficient \citep{ziegler2020finetuning}.
During training, the policy network $\pi_\theta$ and the value network $V_\phi$ will be optimized jointly.
\footnote{In this work, we use the RL4LMs library to learn the response generation policy, so please refer to \citet{ramamurthy2023is} for more algorithm implementation details.}

\begin{table}[t]
    \centering
    \small
    \begin{tabular}{p{0.93\linewidth}}
    \toprule
     \multicolumn{1}{c}{\textbf{In-Context Example}} \\
     \midrule
     \textbf{Instruction}: The following is a conversation with an AI assistant. The assistant is providing an answer based on a knowledge passage. \\
    \textbf{Human}: I recently discovered rap music and I'm so intrigued by it! Do you listen to rap music? \\
    \textbf{Knowledge Text}: Rapping is a musical form of vocal delivery that incorporates ''rhyme, rhythmic speech, and street vernacular'', which is performed or chanted in a variety of ways.\\
    \textbf{AI}: No, I'm a bot and can't hear. I know that it's a form of music that involves chanting and rhythmic speech. \\
     \midrule
     \multicolumn{1}{c}{\textbf{LLM Input}} \\
     \midrule
     \textbf{Human}: Chevrolet is my all time favorite car brand, have you heard of it? \\
     \textbf{Knowledge Text}:  The Chevrolet Chevy II/Nova was a small automobile manufactured by Chevrolet, and produced in five generations for the 1962 through 1979, and 1985 through 1988 model years.\\ 
     \midrule
     \multicolumn{1}{c}{\textbf{LLM Output}} \\
     \midrule
     \textbf{Model Output 1}:  Yes, I'm familiar with Chevrolet. It's a car brand \textcolor{gray}{\textit{that has been around since the early 1900s and }} has produced popular models like the Chevy II/Nova, \textcolor{gray}{\textit{Corvette, Camaro, and Impala}}. \\ 
     \midrule
     \textbf{Model Output 2}: Yes, I heard of Chevrolet. It's a popular car brand that has been manufacturing cars for several decades. Chevy II/Nova was produced in five generations between 1962 and 1988. \\
    \bottomrule
    \end{tabular}
    \caption{An example from the validation set of FaithDial for human expert to do pair-wise comparison between the outputs of text-davinci-003 and GPT-3.5-turbo.
    The \textcolor{gray}{\textit{gray text}} is the unfaithful model-generated response caused by the bias from the LLM. 
    \label{tab:human_demo}}
\end{table}

\subsection{Blending Reward Functions via Expert Demonstrations}
\label{sec:our_reward}
\paragraph{Blending Reward Functions.}
We propose leveraging existing automatic metrics to evaluate the faithfulness and accuracy of generated responses and directly learn a policy that balances the trade-off between these two factors. 
This approach eliminates the need for large-scale human annotations and utilizes the available metrics to guide the learning process effectively. 
By optimizing the policy based on these metrics, the model achieves improved performance in generating faithful and accurate responses.

Formally, the new reward function is defined as:
\begin{equation}
\label{eq:our_reward}
    R_t = \alpha R_t^{\text{acc}}(\vec{u}, \vec{y}) + (1-\alpha)R_t^{\text{faith}}(\vec{u}, K)
\end{equation}
where $R_t^{\text{acc}}$ is the accuracy metric measuring the similarity between the generated response $\vec{u}$ and the ground-truth reference $\vec{y}$, 
$R_t^{\text{faith}}$ is the faithfulness metric evaluating the factual consistency between the generated response $\vec{u}$ and the knowledge text $K$, and $\alpha \in [0, 1]$ is a coefficient used to balance the accuracy and faithfulness of generated responses.
In this work, we choose SacreBLEU \citep{post-2018-call} as $R_t^{\text{acc}}$ and BERTScore \citep{Zhang2020BERTScore} as $R_t^{\text{faith}}$, as they are recognized as effective evaluation metrics in the knowledge-grounded dialogue generation task \citep{10.1162/tacl_a_00529,dziri-etal-2022-evaluating}. 
Note that $R_t^{\text{acc}}$ and $R_t^{\text{faith}}$ are only assigned to the final token in the generated response.
This reward will also be regularized with the token-level KL penalty the same way as in \autoref{eq:ppo_reward}.

\paragraph{Learning $\alpha$ from Expert Demonstrations.}
The selection of the coefficient $\alpha$ is crucial for learning an effective policy.\footnote{Note that we cannot learn the coefficient $\alpha$ during training, because the reward is part of PPO's learning objective.}
Grid search on the validation set is a common strategy but can be computationally intensive and prone to overfitting on the validation set.
To address this issue, we propose to learn the coefficient $\alpha$  from a few number of expert pair-wise comparison demonstrations.
Since the primary requirement of the reward function is to differentiate faithful and accurate responses, leveraging expert demonstrations can provide valuable insights for determining an appropriate value for $\alpha$. 
Additionally, this approach can reduce the computational burden and help ensure the generalizability of the learned policy.

Specifically, we leverage two state-of-the-art large language models (LLMs), text-davinci-003 and GPT-3.5-turbo\footnote{\url{https://platform.openai.com/docs/api-reference}}, to generate 25 responses respectively based on one in-context example.
Then we shuffle the presentation order and ask an NLP expert to do pair-wise comparisons between the two model outputs.
An illustration example is provided in \autoref{tab:human_demo}.
Next, we compute the reward using \autoref{eq:our_reward} for both models' outputs, and align our reward comparison results with the expert pair-wise comparison results.
The alignment is done by iterating values of $\alpha$ and finding the optimal value that maximizes the Pearson correlation coefficient \citep{pearson1895vii} between the expert pair-wise comparison results and our reward pair-wise comparison results.
By learning $\alpha$ from these demonstrations, we can effectively calibrate the balance between faithfulness and accuracy in the generated responses. 
\SetAlCapNameFnt{\small}
\SetAlCapFnt{\small}

\begin{algorithm}[ht]
\small
  \caption{Learning $\alpha$ from expert demonstrations}
  \label{alg:alg}
  \textbf{Input}: 
    LLMs' outputs $\{(\vec{u}_n^1, \vec{u}_n^2\}_{n=1}^N$.
  The expert demonstrations of pair-wise comparison on two LLMs' outputs $\{\hat{p}_n\}_{n=1}^N$.\\
  \textbf{Output}: The optimal blending coefficient $\alpha_{human}$
  \vspace{-3mm}
  \begin{algorithmic}[1] 
    \FOR {$\alpha = 0.00,\dots,1.00 $}
    \STATE Compute our reward on two LLMs' outputs using \autoref{eq:our_reward} and get the scores $\{(r_n^1, r_n^2)\}_{n=1}^N$.
    \STATE Get the pair-wise comparison of our reward $p^{\alpha}_n = \text{argmax}(r_n^1, r_n^2)$, for $n=1,\dots, N$ 
    \STATE Compute the Pearson correlation coefficient $r$ between $\{\hat{p}_n\}_{n=1}^N$ and $\{p^{\alpha}_n\}_{n=1}^N$
    \ENDFOR
    \STATE Save the optimal $\alpha$ which achieves the highest Pearson correlation coefficient $r_{\max}$ as $\alpha_{human}$ 
  \end{algorithmic}
\end{algorithm}

\section{Experiments}
The experiments in this section are designed to answer the following research questions:
\begin{enumerate}[noitemsep,topsep=0pt,leftmargin=*,label=RQ\arabic*]
    \item Does there exist a trade-off between faithfulness and accuracy in the knowledge-grounded dialogue generation task?
    \item Can {the proposed} method improve the faithfulness and accuracy of model-generated responses?
    \item Can expert demonstrations help effectively calibrate the values of $\alpha$ to learn a better policy?
\end{enumerate}

\begin{table*}[t]
  \centering
  \small
  \begin{tabular}{l|l|cc|cc|c}
    \toprule
    & & \multicolumn{2}{c|}{\textbf{Accuracy}} & \multicolumn{2}{c|}{\textbf{Faithfulness}} & \\
    \cmidrule(lr){1-1} \cmidrule(lr){2-2} \cmidrule(lr){3-4} \cmidrule(lr){5-6} \cmidrule(lr){7-7}
    \textbf{Dataset} & \textbf{Method} & \textbf{SacreBLEU} & \textbf{ROUGE-L} & \textbf{BERTScore} & \textbf{Token-F1} & \textbf{Overall} \\
    \midrule
    \multirow{4}{*}{MultiDoc2Dial} & \textbf{R3} & 31.10 & 41.40 & - & - & -  \\
     & \textbf{T5-SFT} & 25.38 & 41.13 & 91.10 & 51.61 & 209.22 \\
     & \textbf{T5-PPO-RoBERTa} & 30.51 & 42.56 &  90.66 & 47.05 & 210.78 \\
     & \textbf{T5-PPO-Ours} & \textbf{31.15} & \textbf{43.28} & \textbf{91.45} & \textbf{51.81} & \textbf{217.69} \\
     \midrule
     \multirow{4}{*}{FaithDial} & \textbf{T5-CTRL} & \textbf{13.75} & 38.57 & 94.42 & 70.91 & 217.65 \\
     & \textbf{T5-SFT} & 13.69 & \textbf{39.58} & 95.13 & 75.49 & 223.89 \\
     & \textbf{T5-PPO-RoBERTa} & 11.87 & 36.75 &  92.06 & 52.05 & 192.73 \\
     & \textbf{T5-PPO-Ours} & 12.75 & 36.98 & \textbf{98.79} & \textbf{94.71} & \textbf{243.23} \\
    \bottomrule
  \end{tabular}
  \caption{\label{tab:auto_result}
  Accuracy and faithfulness evaluation results on the test set of MultiDoc2Dial and FaithDial. 
  The results of \textbf{R3} are reprinted from the original paper \citep{bansal-etal-2022-r3}.
  The results of \textbf{T5-CTRL} come from the code and data released by \citet{10.1162/tacl_a_00529}. The other results come from our implementation.
  The \textbf{Overall} score is a sum of accuracy and faithfulness scores.
  }
\end{table*}

\begin{table*}[t]
  \centering
  \small
  \begin{tabular}{l|c|cc|cc|c}
    \toprule
    & & \multicolumn{2}{c|}{\textbf{Accuracy}} & \multicolumn{2}{c|}{\textbf{Faithfulness}} & \\
    \cmidrule(lr){1-1} \cmidrule(lr){2-2}\cmidrule(lr){3-4} \cmidrule(lr){5-6} \cmidrule(lr){7-7}
    \textbf{Dataset} & $\alpha$ & \textbf{SacreBLEU} & \textbf{ROUGE-L} & \textbf{BERTScore} & \textbf{Token-F1} & \textbf{Overall} \\
    \midrule
    \multirow{4}{*}{MultiDoc2Dial} & $\alpha_{human}$ & \textbf{31.15} & 43.28 & 91.45 & \textbf{51.81} & \textbf{217.69} \\
     & $\alpha=1.00$ & 30.60 & 43.04 & \textbf{91.67} & 51.80 & 217.11 \\
     & $\alpha=0.00$ & 27.45 & \textbf{44.44} & 91.37 & 51.12 & 214.38 \\
     & $\alpha=0.25$ & 29.59 & 44.32 & 91.55 & 52.19 & 217.65 \\
    \midrule
    \multirow{4}{*}{FaithDial} & $\alpha_{human}$ & 12.75 & 36.98 & \textbf{98.79} & \textbf{94.71} & \textbf{243.23} \\
     & $\alpha=1.00$ & \textbf{13.09} & \textbf{38.36} & 95.97 & 81.98 & 229.40 \\
     & $\alpha=0.00$ & 12.59 & 36.95 & 91.57 & 54.47 & 195.58 \\
     & $\alpha=0.85$ & 13.30 & 36.94 & 98.30 & 92.28 & 240.82 \\
    \bottomrule
  \end{tabular}
  \caption{\label{tab:ablation_result}
  Model performances under different values of $\alpha$ on the test set of MultiDoc2Dial and FaithDial. 
  The \textbf{Overall} score is a sum of accuracy and faithfulness scores.
  }
\end{table*}

\subsection{Experimental Setups}
\label{sec:exp_setup}
\paragraph{Benchmark Datasets.}
We choose two information-seeking conversation datasets as our benchmarks: MultiDoc2Dial \citep{feng-etal-2021-multidoc2dial} and FaithDial \citep{10.1162/tacl_a_00529}.
Both datasets contain two participants in each conversation: a user (or seeker) who initiates the conversation with a question, and a system (or wizard) who answers the user's question by referring to a piece of knowledge text.
Each conversation {contains} several turns and {probably} topic shifts.
The datasets statistics are demonstrated in \autoref{tab:dataset}.
\begin{table}[ht]
    \centering
    \small
    \begin{tabular}{l|ccc|c}
    \toprule
     \textbf{Dataset} & \textbf{Train} & \textbf{Valid} & \textbf{Test} & \textbf{Know. Tok.} \\
     \midrule
     MultiDoc2Dial & 21,453 & 4,201 & 4,094 & 106 \\
     FaithDial & 18,357 & 3,417 & 3,539 & 27 \\
     \bottomrule
    \end{tabular}
    \caption{Benchmark datasets statistics, where \textbf{Know. Tok.} indicates the average number of tokens for each input knowledge text. 
    \label{tab:dataset}
    }    
\end{table}

\paragraph{Competitive Methods.}
We compare with the following knowledge-grounded dialogue generation methods:
\begin{itemize}[noitemsep,topsep=0pt,leftmargin=*]
    \item \textbf{R3} \citep{bansal-etal-2022-r3}: a retriever-reranker-reader system that achieves state-of-the-art performance on MultiDoc2Dial. The system uses a bi-encoder DistilSPLADE \citep{formal2021splade} retriever to fetch top-100 relevant knowledge passages from the corpus, then applies a RoBERTa-based \citep{liu2019roberta} cross-encoder to rerank the top-100 knowledge passages, finally passes the top-10 reranked knowledge passages to a T5-based FiD \citep{izacard-grave-2021-leveraging} to generate the response.
    \item \textbf{T5-CTRL} \citep{10.1162/tacl_a_00529}: a controlled generation method that achieves state-of-the-art performance on FaithDial. Following \citet{rashkin-etal-2021-increasing}, it sets control feature tokens based on measures of entailment, lexical precision, and objective voice of the ground-truth response, to steer a T5-base model \citep{raffel2020exploring} generating responses faithful to the input knowledge texts.
    \item \textbf{T5-SFT}: a supervised fine-tuning baseline for both datasets. This method directly fine-tunes the T5-base model with maximum likelihood estimation on ground-truth responses in the full training set. 
    \item \textbf{T5-PPO-RoBERTa}: a baseline reward function for PPO, which is a RoBERTa-base model that discriminates whether the current output is the ground-truth response given the conversation history and knowledge text. The purpose of this reward model is to encourage the model-generated response matching the ground-truth reference. The reward model is trained with an equal number of positive and negative examples, where the negative examples come from the outputs of \textbf{T5-SFT}. For PPO training, we use \textbf{T5-SFT} to initialize the policy network and value network.
    \item \textbf{T5-PPO-Ours}: our proposed reward function for PPO introduced in \autoref{sec:our_reward}. We use \textbf{T5-SFT} to initialize the policy network and value network, and update them jointly with PPO.
\end{itemize}

\begin{table*}[t]
    \centering
    \small
    \begin{tabular}{p{0.15\linewidth}|p{0.75\linewidth}}
    \toprule
    \textbf{History} &
    User: I am aspiring to be an accountant. I want to know more about accounting. \\
    \midrule
    \textbf{Knowledge Text} & \textcolor{teal}{Accounting, which has been called the ''language of business'', measures the results of an organization's economic activities and conveys this information to a variety of users, including investors, creditors, management, and regulators.} \\
    \midrule
    \textbf{Human Reference} & Accounting has been called the ''language of business''. \\
    \midrule
    \textbf{T5-SFT} & Accounting has been called the ''language of business''. It measures the results of an organization's economic activities and conveys this information to a variety of users. \\
    \midrule
    \textbf{T5-PPO-Acc.} ($\alpha=1.00$) & I see, did you know that accounting has been called the ''language of business''? It measures the results of an organization's economic activities and conveys this information to a variety of users? \\
    \midrule
    \textbf{T5-PPO-Faith.} ($\alpha=0.00$) & Ah, accounting is called the language of business. \\
    \midrule
    \textbf{T5-PPO-Ours} ($\alpha_{human}$) & Accounting has been called the ''language of business'', measures the results of an organization's economic activities and conveys this information to a variety of users, including investors, creditors, management, and regulators. \\
    \bottomrule
    \end{tabular}
    \caption{\label{tab:fd_case}
    Different model-generated responses sampled from the test set of FaithDial. 
    The \textcolor{teal}{green text} highlights the ground-truth knowledge span.
    }
\end{table*}

\paragraph{Implementation Details.}
We choose T5-base as our backbone language model for all experiments.
For the input knowledge text in MultiDoc2Dial, we use the ground-truth knowledge passage preprocessed by the official code\footnote{\url{https://github.com/IBM/multidoc2dial/blob/main/scripts/run_data_preprocessing.sh}}, because we focus on the agent response generation subtask.
For \textbf{T5-SFT} baseline, we fine-tune the model on the ground-truth response from the full training set. 
We fine-tune the model for 10 epochs using the AdamW optimizer \citep{loshchilov2018decoupled} with a linear decaying learning rate starting from $1\times10^{-5}$. 
During inference, we use beam search with beam size of 4 to generate the final response.

The RoBERTa-base reward model in \textbf{T5-PPO-RoBERTa} is trained to discriminate the \textbf{T5-SFT} generated responses and the ground-truth responses. 
We train the reward model for 10 epochs using the AdamW optimizer with a constant learning rate of $1\times10^{-6}$.
The reward model achieves 89\% accuracy on the test set of MultiDoc2Dial and 96\% accuracy on the test set of FaithDial.

For learning $\alpha$ from expert demonstrations, we collect 25 pair-wise comparison demonstrations on each dataset respectively.
We find the optimal value of $\alpha_{human}$ on MultiDoc2Dial is 0.04 with a Pearson correlation coefficient of 0.2278, and the optimal value of $\alpha_{human}$ on FaithDial is 0.92 with a Pearson correlation coefficient of 0.2865.

For all PPO experiments, the policy network and value network share the same base model initialized from \textbf{T5-SFT} but separate the last output layer. 
The output layer of the value network is a linear network that maps the last hidden state to a scalar value.
We update the parameters for 10,000 iterations using the AdamW optimizer \citep{loshchilov2018decoupled} with a learning rate of $5\times10^{-7}$.
The policy is evaluated on the full validation set every 100 iterations, and the final policy is the one that achieves the highest total scores in accuracy and faithfulness on the validation set.
We use top-k ($k=50$) sampling to generate trajectories during training, and use beam search with beam size of 4 to generate the final response during testing.

\paragraph{Evaluation Metrics.}
We follow the prior works \citep{feng-etal-2021-multidoc2dial,10.1162/tacl_a_00529} to evaluate the accuracy and faithfulness of model-generated responses.
Specifically, we use SacreBLEU \citep{post-2018-call} and ROUGE-L \citep{lin-2004-rouge} to evaluate the accuracy, which measures the similarity between model-generated responses and ground-truth responses;
we use the F1 measure of BERTScore \citep{Zhang2020BERTScore} and the token-level F1 scores to evaluate the faithfulness, which computes the similarity between model-generated responses and input knowledge texts.

\begin{table*}[t]
    \centering
    \small
    \begin{tabular}{p{0.15\linewidth}|p{0.75\linewidth}}
    \toprule
    \textbf{History} & 
    User: Can those over the age of 18 drive without supervision while holding a learner permit? \\
    \midrule
    \textbf{Knowledge Text} & Learner permit restrictions // \textcolor{teal}{No matter what age you are,  if you hold a learner permit,  you may not drive unless accompanied by a supervising driver age 21 or older who has a valid license to operate the vehicle you are driving.} \underline{For example,  only a} 
    \underline{person with a motorcycle license may supervise a person learning to drive a motorcycle. You m-} \underline{ay not drive with a learner permit  on any street within a park in New York City on any bridge or}
    \underline{tunnel under the jurisdiction of the Tri-borough Bridge and Tunnel Authority on the Cross Coun-} 
    \underline{ty}, Hutchinson River, Saw Mill River, or Taconic State parkways in Westchester County in a DMV road test area. \\
    \midrule
    \textbf{Human Reference} & No. Learner permit holders of any age can't drive without a supervising driver. \\
    \midrule
    \textbf{T5-SFT} & \textcolor{gray}{\textit{Yes, (unfaithful)}} you may not drive unless accompanied by a supervising driver age 21 or older who has a valid license to operate the vehicle you are driving. \textcolor{red}{For example, only a person with a motorcycle license may supervise (incorrect span)} \\
    \midrule
    \textbf{T5-PPO-Acc.} ($\alpha=1.00$) & \textcolor{red}{You may not drive with a learner permit on any street within a park in New York City on any bridge or tunnel under the jurisdiction of the Tri-borough Bridge and Tunnel Authority on the Cross County. (incorrect span)} \\
    \midrule
    \textbf{T5-PPO-Faith.} ($\alpha=0.00$) & \textcolor{gray}{\textit{Yes. (unfaithful)}} \\
    \midrule
    \textbf{T5-PPO-Ours} ($\alpha_{human}$) & No. You may not drive unless accompanied by a supervising driver age 21 or older who has a valid license to operate the vehicle you are driving. \\
    \bottomrule
    \end{tabular}
    \caption{\label{tab:mdd_case}
    Different model-generated responses sampled from the test set of MultiDoc2Dial. 
    The \textcolor{teal}{green text} highlights the ground-truth knowledge span.
    The \textcolor{gray}{\textit{gray text}} is the unfaithful model-generated response caused by the bias from the LLM. 
    The \textcolor{red}{red text} is the incorrect model-generated response misled by the \underline{text} that contains redundant information.
    }
\end{table*}

\subsection{Result Analysis}

\paragraph{RQ1: Faithfulness v.s. Accuracy.}
We first investigate whether there exists a trade-off between faithfulness and accuracy in the knowledge-grounded dialogue generation task.
The empirical results in \autoref{tab:auto_result} indicate the answer depends on the input knowledge text.
When the input knowledge text contains redundant and irrelevant information, as in the case of MultiDoc2Dial, a good method can help improve both accuracy and faithfulness.
Because the ground-truth reference can help locate the relevant text span in a long knowledge passage.

On the other hand, when the input knowledge text only contains the precise information needed for the answer, as in the case of FaithDial, the model may learn to directly copy from the knowledge text, which improves the performance in faithfulness but degrades the performance in accuracy.
In this case, a good method should balance the performance improvement trade-off between faithfulness and accuracy.

\paragraph{RQ2: Our Method v.s. Others.}
Generally speaking, our method achieves the best overall performance compared with other competitive baselines in both datasets, as shown in \autoref{tab:auto_result}.
In MultiDoc2Dial, \textbf{T5-PPO-Ours} achieves the best performance in both accuracy and faithfulness evaluation, indicating our learning method has the advantage of locating the relevant information from a long knowledge document.
\textbf{T5-PPO-RoBERTa} achieves improved performance in accuracy but degraded performance in faithfulness compared with \textbf{T5-SFT}, because its reward function is designed to only encourage the model-generated response matching the ground-truth reference.

In FaithDial, \textbf{T5-PPO-Ours} achieves the best performance in faithfulness but degraded performance in accuracy compared with \textbf{T5-SFT}.
As discussed in \textbf{RQ1}, the knowledge text of FaithDial contains the exact relevant information for the answer, consequently, the model learns to directly copy the knowledge text as its response when further trained with PPO.
A generation example is provided in \autoref{tab:fd_case}. 
Besides, \textbf{T5-PPO-RoBERTa} achieves the worst performance even if its reward model has 96\% test accuracy, which indicates the reward model also learns some shortcut features in model-generated responses (e.g. the n-gram overlap between response and knowledge text).

\paragraph{RQ3: Calibrating $\alpha$.}
As illustrated in \autoref{tab:ablation_result}, we experiment with different values of $\alpha$ for our reward function, and find that the value learned from expert demonstrations $\alpha_{human}$ achieves the best overall performance in both datasets. 
Single metric, i.e. $\alpha=0.00$ and $\alpha=1.00$, only achieves good performance in a single evaluation dimension, and still generates inaccurate or unfaithful responses, as shown in \autoref{tab:mdd_case}.

Additionally, we compared $\alpha_{human}$ with values found by grid search which achieves the highest total score on the validation set, i.e. $\alpha=0.25$ and $\alpha=0.85$.
Surprisingly, we find $\alpha_{human}$ achieves better overall performance in both datasets, which indicates few expert demonstrations can not only help calibrate the values to learn a better policy, but also provide a good generalization ability for the reward function.
\section{Conclusion}
This work investigates how to improve faithfulness and accuracy in knowledge-grounded dialogue generation tasks.
Firstly, we identify three major sources for the problem of generating unfaithful and inaccurate responses: the bias in LLMs, the irrelevant information in knowledge texts, and the characteristics of the supervised learning algorithm.
Then, we solve the problem by applying a reinforcement learning algorithm with a novel reward function.
Our reward function can be used as a cost-effective approximation to the human preference reward model learned from massive high-quality human preference annotations.
Finally, we validate the effectiveness of our method in two information-seeking conversation datasets.
The empirical experiment results show that our method can outperform other strong supervised learning baselines.
\section*{Limitations}

The good performance of PPO algorithm not only requires a good reward function, but also relies on a good initial policy.
The initial policy ensures the trajectories sampled from the current policy are of high quality, which benefits the convergence to a better policy.
Therefore, the pretraining of better language models is always helpful for this method.
In addition, learning on-policy RL algorithms requires large GPU memory, so more memory-efficient training or model compression techniques can further benefit the RL training.

\section*{Ethics Statement}
This work complies with the ACL Ethics Policy. 
Both benchmark datasets and baseline models are collected from public academic resources, and do not contain harmful, unfair, or discriminating content.
We will also make our data and code open-sourced once this paper is made public, in order to provide easily reproducible experiment configurations for future research.

\bibliography{anthology,custom}
\bibliographystyle{acl_natbib}


\end{document}